\documentclass[a4paper,fleqn]{cas-dc}

\usepackage[numbers]{natbib}
\usepackage{multirow}
\usepackage{subfig}
\usepackage{wrapfig}
\usepackage{balance}
\usepackage{threeparttable}
\def\tsc#1{\csdef{#1}{\textsc{\lowercase{#1}}\xspace}}
\tsc{WGM}
\tsc{QE}

\begin{document}
\let\WriteBookmarks\relax
\def\floatpagepagefraction{1}
\def\textpagefraction{.001}
\let\printorcid\relax

\shorttitle{A novel emotion recognition network in textual conversation based on curriculum learning strategy}

\shortauthors{Li et al.}  

\title[mode = title]{ERNetCL: A novel emotion recognition network in textual conversation based on curriculum learning strategy}

\tnotetext[1]{Accepted by Knowledge-Based Systems (KBS). DOI: 10.1016/j.knosys.2024.111434. Received 11 September 2023, Revised 13 December 2023, Accepted 21 January 2024. \textit{E-mail address:} lijfrank@hust.edu.cn (J. Li), wangxiaoping@hust.edu.cn (X. Wang).}

\author[1,2,3,4]{Jiang Li}
\cormark[1]
\author[1,3,4]{Xiaoping Wang}
\cormark[1]
\cortext[1]{Corresponding author at: School of Artificial Intelligence and Automation, Huazhong University of Science and Technology, Wuhan 430074, China.}
\author[1,3,4]{Yingjian Liu}
\author[1,3,4]{Zhigang Zeng}
\address[1]{School of Artificial Intelligence and Automation, Huazhong University of Science and Technology, Wuhan 430074, China}
\address[2]{Institute of Artificial Intelligence, Huazhong University of Science and Technology, Wuhan 430074, China}
\address[3]{Hubei Key Laboratory of Brain-inspired Intelligent Systems, Huazhong University of Science and Technology, Wuhan 430074, China}
\address[4]{Key Laboratory of Image Processing and Intelligent Control (Huazhong University of Science and Technology), Ministry of Education, Wuhan 430074, China}

\begin{abstract}
Emotion recognition in conversation (ERC) has emerged as a research hotspot in domains such as conversational robots and question-answer systems. How to efficiently and adequately retrieve contextual emotional cues has been one of the key challenges in the ERC task. Existing efforts do not fully model the context and employ complex network structures, resulting in limited performance gains. In this paper, we propose a novel emotion recognition network based on curriculum learning strategy (ERNetCL). The proposed ERNetCL primarily consists of temporal encoder (TE), spatial encoder (SE), and curriculum learning (CL) loss. We utilize TE and SE to combine the strengths of previous methods in a simplistic manner to efficiently capture temporal and spatial contextual information in the conversation. To ease the harmful influence resulting from emotion shift and simulate the way humans learn curriculum from easy to hard, we apply the idea of CL to the ERC task to progressively optimize the network parameters. At the beginning of training, we assign lower learning weights to difficult samples. As the epoch increases, the learning weights for these samples are gradually raised. Extensive experiments on four datasets exhibit that our proposed method is effective and dramatically beats other baseline models.
\end{abstract}

\maketitle

\section{Introduction}
\begin{figure}[htbp]
    \centering
    \includegraphics[width=3.3in]{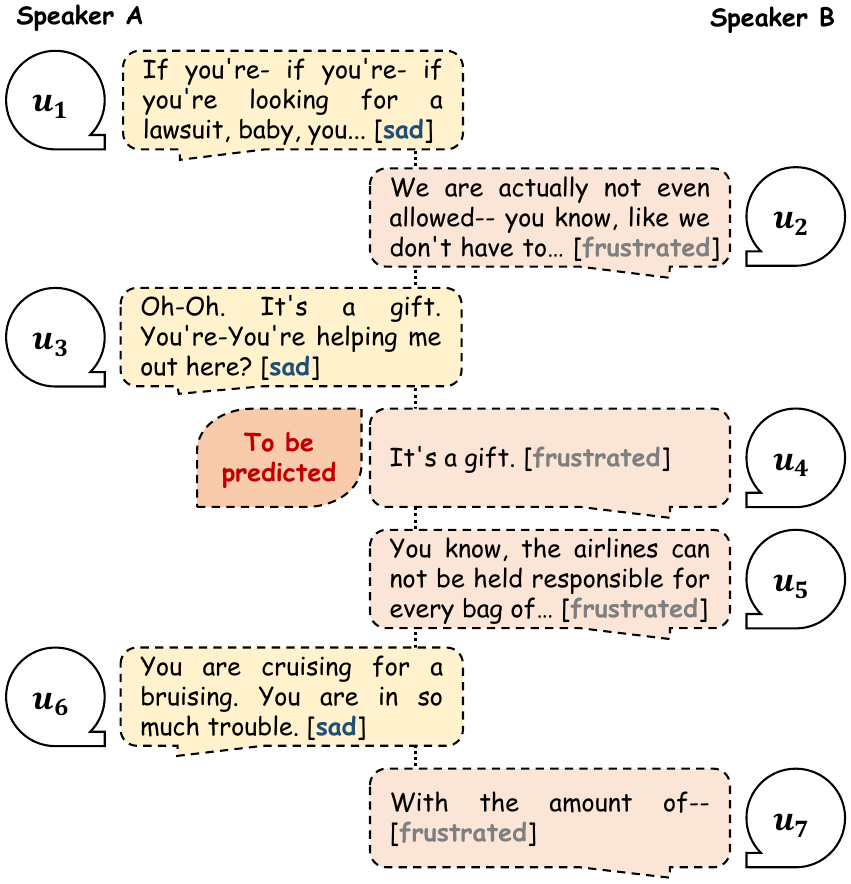}
    \caption{A conversational scenario. Combining current and contextual information is needed to comprehensively determine the emotion of the utterance to be predicted.}
    \label{fig:conversation}
\end{figure}

Emotional dialogue system strives to enable chatbots to recognize, understand, and express emotions when interacting with users, yielding more attractive and various reactions. Since emotional dialogue system empowers robots to interact with users in a more empathetic and human-like fashion, it has the potential to become a hot research field. Emotion is regarded as a brief but intense physiological response of the brain to a stimulus, and it is expressed in the form of facial expression, vocal tone, and behavioral change~\cite[]{jia2022beyond}. Enabling robots to identify and understand emotions is a key aspect of realizing an emotional dialogue system. Hence, emotion recognition in conversation (ERC) has become a centerpiece technology to augment the performance of many applications and has received much attention in realms such as question-answer systems~\cite[]{du2023sentiment}, opinion mining~\cite[]{Venugopalan2022Anenhanced}, recommender systems~\cite[]{polignano2021towards}, and conversational robots~\cite[]{Singh2022Knowing}. The target of ERC is to automatically recognize the emotions of the participants as they say each utterance according to the conversational content. The modeling style of ERC is dissimilar to that of non-conversational emotion recognition due to its multi-turn conversational scenario and the presence of natural emotion transition~\cite[]{wang2020relational,chenqian2020relation}. As shown in Figure~\ref{fig:conversation}, it is extremely hard for the ERC system to accurately derive the corresponding emotional state based on the current utterance alone, but rather, it needs to make a comprehensive judgment by combining the contextual information in the conversation. Accordingly, how to maximally model the context in the conversation is of paramount importance to the ERC task.

There are a number of ERC efforts based on contextual modeling. Depending on the network structure of these works, they are primarily categorized into recurrent neural network (RNN)-based methods, graph neural network (GNN)-based methods, and multi-head attention (MHA)-based methods. RNN-based models typically construct different modules utilizing long short-term memory (LSTM) or gated recurrent unit (GRU) to capture emotional cues at different aspects. Ghosal et al.~\cite{ghosal2020cosmic} utilized different GRUs to establish five distinct states in a dialogue while adding commonsense knowledge to improve performance. Hu et al.~\cite{hu2021dialoguecrn} drew inspiration from the emotion cognition theory and constructed LSTM-based multi-turn inference module to extract emotional cues. These methods tend to extract limited information from the nearest neighbors, causing unsatisfactory performance improvements. GNN-based models construct the conversation as a graph from the spatial perspective to extract emotional cues between utterances. Lee et al.~\cite{leechoi2021graph} considered the ERC task as a dialogue-based relation extraction and constructed a heterogeneous graph to capture the relationships between arguments in the dialogue. Li et al.~\cite{li2021pastpresent} presented a psychological-knowledge-aware interaction graph by constructing a locally connected graph and introducing different types of edges to simulate the psychological interaction between utterances. These approaches can capture long-distance contextual information, thus effectively mitigating the shortcomings of RNN-based methods. Analogous to GNN-based models, MHA-based ones exploit global attention to retrieve long-distance emotional cues. Indeed, MHA is a network structure that operates graph attention~\cite[]{velickovic2018graph} on the fully connected graph, i.e., MHA can be regarded as a special type of GNNs. HiTrans~\cite[]{li2020hitrans} employed two separate Transformer networks to model a conversation, where the low-level Transformer was utilized to capture local utterance information, and the high-level Transformer was leveraged to extract global contextual information. CoG-BART~\cite[]{li2022contrast} modeled the conversation with the powerful comprehension and generation capabilities of BART~\cite[]{lewis2020bart} while employing contrastive learning to further upgrade performance.

Although GNN- and MHA-based approaches address the drawback of the inability to capture long-distance contextual information, they ignore temporal sequence information in the conversation. To summarize, most of the existing efforts often fail to model the conversation from both temporal and spatial perspectives, i.e., they do not take into account the combination of temporal and spatial contextual information, making them infeasible to adequately extract contextual information. On the other side, the network structure of most ERC works is overly complex, e.g., employing encoder-decoder architecture~\cite[]{li2020hitrans,li2022contrast}, containing too many components~\cite[]{hu2021dialoguecrn,leechoi2021graph}, and integrating commonsense knowledge~\cite[]{ghosal2020cosmic,li2021pastpresent}, causing limited performance improvements to be obtained. This suggests that some complex structures may be redundant and useless in current ERC models. We try to leverage GRU and MHA networks in a straightforward manner to achieve a combination of RNN- and MHA-based approaches, which in turn can effectively extract spatio-temporal contextual information.

Previous studies~\cite[]{ghosal2020cosmic,shen2021directed,shen2021dialogxl} have demonstrated that emotion shift is one of reasons plaguing the failure of the ERC system. The emotion-shift problem refers to the difficulty for the ERC system to effectively handle scenarios where the utterances are consecutive but their true emotions are different. Theoretically, the higher the frequency of emotion shifts that occur in a conversation, the larger the difficulty of that conversation, i.e., the more difficult it is to classify samples (i.e., utterances) in that conversation. To mitigate the negative effects caused by emotion shift, we introduce a curriculum learning (CL)~\cite{bengio2009curriculum} strategy in the ERC task. Firstly, we adopt the frequency of emotion shifts in the conversation to measure the difficulty score of CL. Then, lower learning weights are assigned to these difficult samples at the beginning of training. As the increase of epoch, the learning weights for these samples are gradually grown. The main thought behind using CL is to first train the model with simpler data subsets, and then gradually employ more difficult subsets until the entire dataset is fully exploited to obtain an optimal model.

In a nutshell, we propose a novel \textbf{E}motion \textbf{R}ecognition \textbf{Net}work in textual conversation based on the \textbf{C}urriculum \textbf{L}earning strategy (ERNetCL). The proposed method not only thoroughly models a conversation from both temporal and spatial perspectives but also alleviates the emotion-shift problem by utilizing the idea of ``training from simple to difficult data" in CL. We conduct extensive comparison and ablation experiments on four public emotion datasets, and the results validate the effectiveness of our ERNetCL. Our contributions are summarized below:
\begin{itemize}
    \item A novel conversational emotion recognition network (i.e., ERNetCL) based on curriculum learning is proposed. ERNetCL combines RNN- and MHA-based methods in a simplistic fashion to capture temporal and spatial contextual information.
    \item We apply curriculum learning strategy to the ERC task in order to progressively optimize the network parameters of ERNetCL from easy to hard. The frequency of emotion shift is used to measure the difficulty score of curriculum learning.
    \item We perform extensive comparative and ablative experiments on four baseline datasets, and the empirical results confirm that our proposed ERNetCL is superior to other baselines.
\end{itemize}

The rest of this paper is structured as follows. In Section~\ref{sec:related}, we describe related works. Section~\ref{sec:ernetcl} corresponds to the methodology of this work, i.e., ERNetCL. Sections~\ref{sec:setup} is the description of the experimental setup. In Section~\ref{sec:result}, we report, discuss, and analyze experimental results. Section~\ref{sec:conclusion} is the summary of this work.

\section{Related work}\label{sec:related}
\subsection{Emotion recognition in conversation}
The goal of emotion recognition in conversation (ERC) is to assign a specific emotion to each utterance based on the content of the conversation. ERC has emerged as an influential research issue owing to its widespread implementation in numerous scenarios such as conversational robots~\cite[]{mensio2018therise} and mental health services~\cite[]{fei2020deep}. The ERC mission differs from non-conversational emotion recognition~\cite[]{wang2020relational,chenqian2020relation} with isolated utterances in that it needs to integrate conversational intent and context. Existing ERC models can be categorized into three groups: recurrent neural network (RNN) based approaches, graph neural network (GNN) based approaches, and multi-head attention (MHA) based approaches. 
Gan et al.~\cite{gan2022dhfnet} proposed a hierarchical feature interactive fusion network that learns the cross impact of conversational emotion recognition and conversational intent recognition through collaborative attention to obtain deep semantic information. Jiao et al.~\cite{jiao2020real} proposed a GRU-based ERC model in which a hierarchical memory network was utilized to extract the interactive information between historical utterances, and a bidirectional GRU was employed to summarize the recent and long-term memory attention weights. Zhao et al.~\cite{zhao2022cauain} explicitly modeled intra- and inter-speaker dependencies by introducing commonsense knowledge as a cue for emotional cause detection in conversations. Bao et al.~\cite{Bao2023SpeakerGuided} presented a speaker-guided encoder-decoder framework called SGED to leverage speaker information for emotion decoding. These methods tend to focus on the near contexts and ignore the long-distance contexts, leading to limited performance improvements of models. 

To address this issue, numerous models based on GNN and MHA have been proposed. Ghosal et al.~\cite{ghosal2019dialoguegcn} proposed the first ERC model based on GNN, which utilized utterances and their associations in a conversation to construct a graph. Shen et al.~\cite{shen2021directed} modeled the dialogue by constructing the directed graph for the task of emotion recognition. Ren et al.~\cite{ren2022lrgcn} utilized a relationship graph network to integrate contextual information and speaker dependencies, and then extracted potential associations between utterances with the help of a multi-attention mechanism. Yang et al.~\cite{yang2023ClusterLevel} utilized pre-trained knowledge adapters to integrate linguistic and factual knowledge, and meanwhile they simplified the high-dimensional supervised contrastive learning space into a three-dimensional affective representation space. Zhong et al.~\cite{zhong2019knowledge} implemented ERC using a hierarchical Transformer, where utterance-level self-attention and context-level self-attention modules were utilized to calculate utterance and context representations, respectively. Song et al.~\cite{song2022supervised} employed the supervised prototypical contrastive learning to strengthen the classification ability of the model while designing a metric function based on inter-class distance to mitigate the effect of extreme samples. Son et al.~\cite{son2022grasp} proposed a relational semantic based prompt guidance model that utilized the large pre-trained model to compensate for the low information density of multi-person conversation. Yang et al.~\cite{yang2023Disentangled} proposed an auxiliary task for target utterance reconstruction based on the variational auto-encoder to improve model performance and regularize the latent spaces.

Although ERC models based on GNN and MHA can capture long-distance contextual information, they are prone to ignore the temporal sequence information in the conversation. Conclusively, most extant ERC methods do not adequately mine contextual emotional cues, and they contain complex network structures. In our work, we attempt to utilize the advantages of RNN and MHA to fully extract contextual information from both temporal and spatial perspectives.

\subsection{Curriculum learning strategy}
Curriculum learning (CL) is a training strategy that mimics the way humans learn curriculum step by step from easy to difficult. The basic idea of CL is to train a model using simpler data subsets, and then gradually adopt harder subsets until an optimal model is obtained. Bengio et al.~\cite{bengio2009curriculum} first proposed the idea of CL under the background of machine learning, which was then widely applied in various fields such as image classification~\cite[]{dogan2020label,cascantebonilla2020curriculum}, object detection~\cite[]{wang2018weakly,soviany2021curriculum}, and speech processing~\cite[]{zhang2019automatic,lotfian2019curriculum} due to its effectiveness. The benefits of applying CL to these realms are mainly summarized as improving the performance of the target task and accelerating the training process~\cite[]{wang2022asurveyoncurriculum}.

Dogan et al.~\cite{dogan2020label} proposed a label similarity curriculum method for image classification. This approach did not employ true labels to train the model in the early stages of the learning process, instead using the probability distribution of classes. Cascante-Bonilla et al.~\cite{cascantebonilla2020curriculum} presented a curriculum labeling method that enhanced the process of selecting the correct pseudo-labels through the use of curriculum based on the extreme value theory. Wang et al.~\cite{wang2018weakly} introduced an easy-to-hard method for weakly- and semi-supervised object detection that fine-tuned weakly annotated images. For unsupervised cross-domain object detection, Soviany et al.~\cite{soviany2021curriculum} utilized a self-paced curriculum approach. They proposed a multi-stage technique for better knowledge transfer from the source domain to target domain. Zhang et al.~\cite{zhang2019automatic} presented a teacher-student strategy for digital modulation classification, which trained the teacher network with feedback from a pre-initialized student network. Lotfian et al.~\cite{lotfian2019curriculum} utilized CL for speech emotion detection, which applied an easy-to-hard batch technique and fine-tuned the learning rate for each bin.

CL has also seen widespread application in natural language processing. Wang et al.~\cite{wang2019dynamically} proposed a joint curriculum technique for neural machine translation that incorporated two levels of heuristics to build domain and denoising curriculums. Zhan et al.~\cite{zhan2021metacl} provided a meta-CL strategy for tackling neural machine translation issue in cross-domain settings. Liu et al.~\cite{liu2020norm} introduced a norm-based CL method for increasing the training efficiency of neural machine translation system. With the use of the word embedding norm, this method measured the difficulty of the sentence, the ability of the model, and the weight of the sentence. Zhou et al.~\cite{zhou2020uncertainty} established an easy-to-difficult ranking by using data-level uncertainty and employed model-level uncertainty to determine the correct timing for augmenting the training set. Tay et al.~\cite{tay2019simple} offered a generative curriculum pre-training strategy for tackling reading comprehension of long narratives. Shen et al.~\cite{shenfeng2020cdl} suggested a curriculum dual learning model, which extended emotion-controlled response generation to a dual task for alternately generating emotional responses and emotional queries.

To the best of our knowledge, only Yang et al.~\cite{yang2022hybrid} and Song et al.~\cite{song2022supervised} have applied CL to ERC tasks. Yang et al.~\cite{yang2022hybrid} proposed a CL framework for ERC that progressively enhanced the ability to recognize emotion through conversation-level and utterance-level curriculums. Song et al.~\cite{song2022supervised} designed a difficulty metric function based on the distance between classes and introduced CL to mitigate the effects of extreme samples. In order to complement the research in this area, we design a new CL scheme for the ERC mission.

\section{Proposed ERNetCL}\label{sec:ernetcl}
\begin{figure*}[htbp]
    \centering
    \includegraphics[width=6.8in]{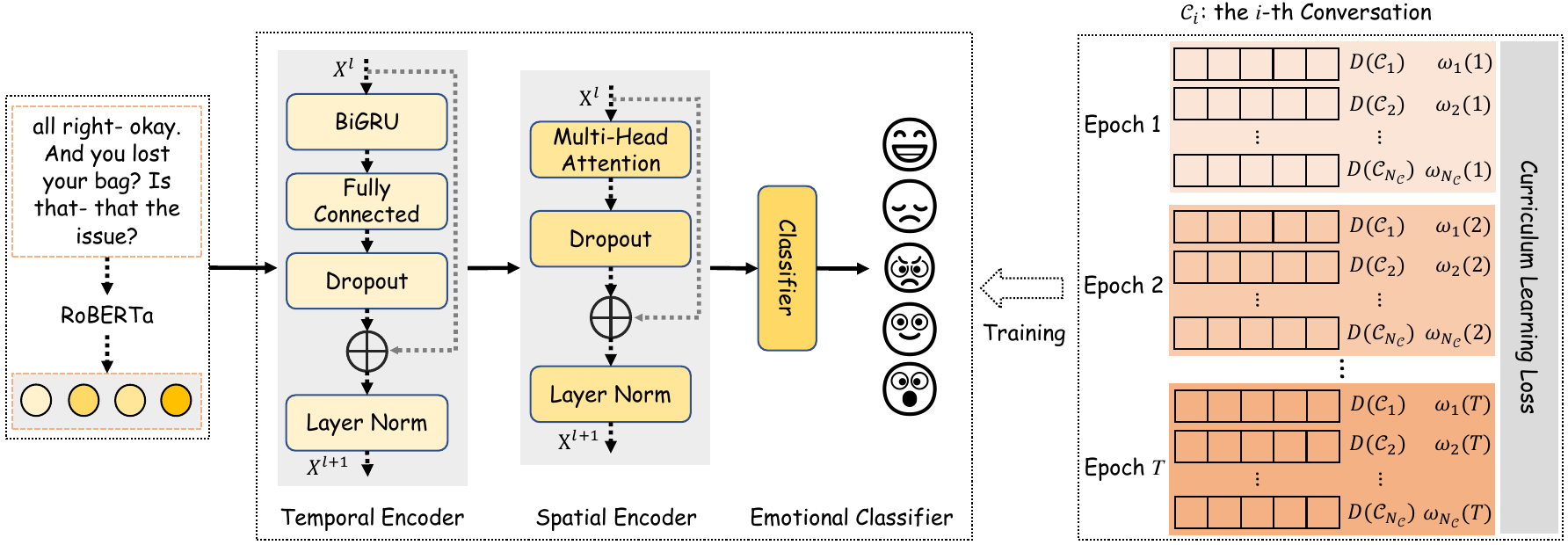}
    \caption{The overall architecture of our proposed ERNetCL. The proposed method sequentially abstracts temporal and spatial contextual cues through temporal and spatial encoders. In the training phase, the curriculum learning loss is adopted to optimize the network parameters instead of the original loss.}
    \label{fig:overall}
\end{figure*}

Modeling the contexts from both temporal and spatial perspectives is conducive to extracting richer emotional information. To achieve this purpose, we adopt GRU- and MHA-based modules in ERNetCL to extract temporal and spatial information, respectively. The overall architecture of the proposed ERNetCL is shown in Figure~\ref{fig:overall}, which mainly includes temporal encoder (TE), spatial encoder (SE), and Emotional Classifier. First, we regard the conversation as a temporal sequence and utilize a GRU-based network, i.e., TE, to extract temporal contextual information; second, to enrich the emotional expressions of utterances from the spatial perspective, an MHA-based network, i.e., SE, is used to capture spatial contextual information in the conversation; and lastly, the features encoded sequentially by TE and SE are applied to emotion classification. Furthermore, in order to simulate the way humans learn curriculum from easy to hard, we replace the original loss with a curriculum learning one to progressively optimize the network parameters of ERNetCL. 

\subsection{Problem definition}
Before introducing the model, let us define the ERC task. Suppose a conversation $\mathcal{C}$ contains $n$ utterances, i.e., $\mathcal{C}=\{c_1,c_2,\cdots,c_n\}$. The goal of ERC is to predict the corresponding emotion $e_i$ based on an utterance $c_i$. If we notate $\mathcal{E}$ as the set of emotion labels, then $e_i$ belongs to $\mathcal{E}$. The number of elements in $\mathcal{E}$ depending on the dataset. For example, in the IEMOCAP dataset, the number of elements is 6, while in the MELD dataset, that is 7. 

The definitions related to emotion shift are as follows. If two consecutive utterances to be determined have the same emotion, then it indicates that there is no emotion shift in these two utterances; otherwise, it indicates that there is an emotion shift in them. In our work, we take into account the speaker identity to determine the number of speaker-specific emotion shifts. More specifically, in a conversation, the utterances said by the same speaker are first formed into a sequence chronologically, and then the number of emotion shifts for that speaker is recorded.

\subsection{Temporal encoder}
Existing efforts~\cite[]{ghosal2020cosmic,hu2021dialoguecrn,zhao2022cauain} have evidenced that the emotional state of the current utterance is affected by its context. To capture the contextual information in the conversation, we adopt a GRU-based temporal encoder (TE) to model that conversation. Concretely, the conversation is first considered as a temporal sequence, where each element of the sequence is an utterance; then, the GRU network, fully connected layer, dropout operation, residual operation, and normalization operation are sequentially employed to extract temporal contextual information. The temporal encoder in Figure~\ref{fig:overall} depicts the above process. Our TE can be expressed by the following equations:
\begin{equation}
    \label{eq:te}
    \begin{split}
    &X_G^l = \mathtt{BiGRU}(X^l),\\
    &X_F^l = \mathtt{DP}(W_F(X_G^l)),\\
    &X^{l+1} = \mathtt{LN}(X^l+X_F^l),
    \end{split}
\end{equation}
where $X^l$ denotes the $l$-th layer feature matrix of utterances in TE; $\mathtt{BiGRU}(\cdot)$ denotes the single-layer bidirectional GRU; $W_F$ is the learnable parameter, which is used to reduce the feature dimension for $X_G^l$ to half of the original one; $\mathtt{DP}(\cdot)$ and $\mathtt{LN}(\cdot)$ denote the dropout and normalization operations, respectively. Note that the residual operation sums the input of BiGRU and the output of dropout operation, and our normalization operation is layer normalization.

\subsection{Spatial encoder}
It is not enough to only consider the contextual information in a conversation from the temporal perspective. Existing studies~\cite[]{vaswani2017attention,dosovitskiy2021an} have demonstrated that the MHA network has a powerful capability for feature extraction. In this subsection, we exploit an MHA-based spatial encoder (SE) to extract spatial contextual information. Specifically, we first assume that the current utterance is directly associated with all utterances in the conversation, i.e., all contextual utterances are regarded as direct neighbors of the current utterance; then, the MHA network, dropout operation, residual operation, and normalization operation are applied to compute spatial contextual information. The above process is depicted by the spatial encoder in Figure~\ref{fig:overall}. Our SE can be formulated as follows:
\begin{equation}
    \label{eq:se}
    \begin{split}
    &\mathrm{X}_A^l = \mathtt{DP}(\mathtt{MHA}(\mathrm{X}^l)),\\
    &\mathrm{X}^{l+1} = \mathtt{LN}(\mathrm{X}^l+\mathrm{X}_A^l),
    \end{split}
\end{equation}
where $\mathrm{X}^l$ denotes the $l$-th layer feature matrix of utterances in SE; similar to TE, $\mathtt{DP}(\cdot)$ and $\mathtt{LN}(\cdot)$ are the dropout and normalization operations, respectively; The residual operation performs a summation between the input of MHA and the output of dropout operation; $\mathtt{MHA}(\cdot)$ denotes the multi-head attention network, it can be expressed as follows,
\begin{equation}
    \label{eq:mha}
    \begin{split}
    &\mathtt{MHA}(\mathrm{X}^l) = W_C^l\mathtt{CAT}(\mathrm{head}_1^l,\mathrm{head}_2^l,\cdots,\mathrm{head}_H^l),\\
    &{s.t.}\ \mathrm{head}_i^l = \mathtt{SoftMax}(\frac{Q_{i}^l \mathrm{X}^l \cdot (K_{i}^l \mathrm{X}^l)^\top}{\cdot \sqrt{d_k}}) V_{i}^l \mathrm{X}^l, 
    \end{split}
\end{equation}
where $W_C^l$, $Q_{i}^l$, $K_{i}^l$, and $V_{i}^l$ denote the trainable parameters; $\mathrm{head}_i^l$ is the $l$-th layer output of the $i$-th head attention network; $H$ is the number of head in MHA; $d_k$ denotes the dimension of $K_{i}^l \mathrm{X}^l$ or $V_{i}^l \mathrm{X}^l$; $\mathtt{CAT}(\cdot)$ denotes the concatenation operation, and $\mathtt{SoftMax}(\cdot)$ denotes the softmax function. It is worth noting that we do not incorporate positional encoding when extracting spatial context information since TE contains temporal sequence information.

\subsection{Emotional classifier}
After multiple layers of temporal and spatial encoding, we can obtain the feature matrix $\mathrm{H}$ of utterances. Finally, the feature dimension of the utterance is reduced to $|\mathcal{E}|$ by the fully connected layer, which in turn yields the predicted emotion $e^{\prime}_i$. The process can be formulated as follows:
\begin{equation}
    \label{eq:classifier}
    \begin{split}
    &y^{\prime}_i = \mathtt{SoftMax}(W_{S} \mathrm{h}_i),\\
    &e^{\prime}_i = \mathtt{ArgMax}(y^{\prime}_i[k]),
    \end{split}
\end{equation}
where $\mathrm{h}_i$ is the feature representation of the $i$-th utterance, $\mathrm{h}_i \in \mathrm{H}$; $W_{S}$ is the learnable parameter, and $\mathtt{ArgMax}(\cdot)$ is the argmax function. 
To optimize the parameters of the model, the loss function is defined as:
\begin{equation}
    \label{eq:loss_}
    \mathcal{L} = - \frac {1}{\sum_{o=1}^{N_{\mathcal{C}}} n(o)} \sum_{i=1}^{N_{\mathcal{C}}}\sum_{j=1}^{n(i)} y_{i,j} \log y^{\prime}_{i,j},
\end{equation}
where $n(i)$ is the number of utterances in the $i$-th conversation, and $N_{\mathcal{C}}$ is the number of all conversations in training set; $y^{\prime}_{i,j}$ denotes the probability distribution of predicted emotion label of the $j$-th utterance in the $i$-th conversation, and $y_{i,j}$ denotes the ground truth label. 

\subsection{Curriculum learning loss}
Several studies~\cite[]{ghosal2020cosmic,shen2021dialogxl,shen2021directed} have revealed that the ERC task suffers from emotion-shift problem. If the number of emotion shifts in a conversation is higher, the probability that utterances in that conversation are correctly classified is lower. To put it differently, the higher the frequency of emotion shifts in a conversation, the greater the classification difficulty of that conversation. Since the architectural design of existing neural networks draws inspiration from the human brain, their training process should also be inspired by the way humans learn. Curriculum learning (CL) is developed precisely with the goal of mimicking the way humans learn curriculum in an easy-to-hard sequence. To alleviate the emotion-shift problem, we adopt the idea of CL to progressively optimize the network parameters of ERNetCL from simple to difficult.

One of the pivotal issues in CL is how to measure difficulty score. Based on the previous analysis, we utilize the frequency of emotion shifts in each conversation to measure the classification difficulty of that conversation (i.e., difficulty score of CL). Our difficulty score is defined as follows:
\begin{equation}
    \label{eq:diff_score}
    D(\mathcal{C}_i) = \frac{1}{N_{sp}(\mathcal{C}_i)} \sum_{s = 1}^{N_{sp}(\mathcal{C}_i)} \frac{N_{sh}(s,\mathcal{C}_i)}{N_{ut}(s,\mathcal{C}_i)},
\end{equation}
where $D(\mathcal{C}_i)$ takes the value range of $[0,1]$; $N_{sp}(\mathcal{C}_i)$ denotes the number of speakers in conversation $\mathcal{C}_i$, $N_{sh}(s,\mathcal{C}_i)$ and $N_{ut}(s,\mathcal{C}_i)$ denote the number of emotion shifts for speaker $s$ and the total number of utterances uttered by $s$ in conversation $\mathcal{C}_i$, respectively. The implication of the above equation is that the higher the average frequency of emotion shifts per participant in a conversation, the greater the classification difficulty of utterances.

We construct a weight function $\omega_{i}(t)$ that varies with the epoch $t$ and classification difficulty $D(\mathcal{C}_i)$. The objective of $\omega_{i}(t)$ is to assign smaller learning weights to difficult utterances (also known as difficult samples) at the beginning of training. As the epoch increases, the learning weights of difficult samples are gradually increased. Consequently, $\omega_{i}(t)$ is defined as follows:
\begin{equation}
    \label{eq:weight_func}
    \begin{split}
    &\omega_{i}(t) = \mathtt{Sigmoid}\left (\frac{R(t)-D(\mathcal{C}_i)}{\sigma}\right ),\\
    &s.t. \ R(t) = \frac{t}{\delta \cdot T}, 
    \end{split}
\end{equation}
where $R(t)$ denotes the ratio of epoch, and $\mathtt{Sigmoid}(\cdot)$ is the sigmoid function; $t$ and $T$ are the current epoch and maximum epoch, respectively; $\sigma$ controls how fast or slow the weight function $\omega_{i}(t)$ grows, and $\sigma \in [0,1]$; and $\delta$ adjusts the size of epoch ratio, and $\delta \geq 1$. 

Drawing on the idea of CL, we define the $t$-th epoch loss function as follows:
\begin{equation}
    \label{eq:loss}
    \mathcal{L}(t) = - \frac {1}{\sum_{o=1}^{N_{\mathcal{C}}} n(o)} \sum_{i=1}^{N_{\mathcal{C}}}\sum_{j=1}^{n(i)} \omega_{i,j}(t) \cdot y_{i,j} \log y^{\prime}_{i,j}(t),
\end{equation}
where $\omega_{i,1}(t)=\omega_{i,2}(t)=\cdots =\omega_{i,n(i)}(t)$, and $t$ stands for the current epoch; as with the original loss, $n(i)$ indicates the number of utterances in the $i$-th conversation, and $N_{\mathcal{C}}$ represents the total number of conversations in training set; $y^{\prime}_{i,j}(t)$ denotes the $t$-th epoch probability distribution of the $j$-th utterance in the $i$-th conversation, and $y_{i,j}$ is the true label. In the training phase, we employ the curriculum learning loss $\mathcal{L}(t)$ instead of the original loss $\mathcal{L}$.

\section{Experimental setup}\label{sec:setup}
\subsection{Datasets}
We evaluate our ERNetCL on four public datasets, including MELD~\cite[]{poria2018meld}, IEMOCAP~\cite[]{busso2008iemocap}, EmoryNLP~\cite[]{zahiri2018emotion}, and DailyDialog~\cite[]{li2017dailydialog}. 

\textbf{MELD} contains 1,433 dialogues with a total of 13,708 utterances. The dataset is composed of multi-party conversation videos sourced from the TV series \textit{Friends}. Each utterance in the dataset is annotated with one of seven emotions, namely \textit{joy}, \textit{anger}, \textit{fear}, \textit{disgust}, \textit{sadness}, \textit{surprise}, and \textit{neutral}. 

\textbf{IEMOCAP} contains 151 conversations with a total of 7,433 utterances. The dataset comprises dyadic conversation videos with ten distinct speakers, with the first eight speakers included in the training set and the remaining two included in the test set. The utterance is labeled with one of six emotions, namely \textit{happy}, \textit{sad}, \textit{neutral}, \textit{angry}, \textit{excited}, and \textit{frustrated}. 

\textbf{EmoryNLP} contains 827 dialogues with a total of 9,489 utterances. The dataset is another one collected from the TV series \textit{Friends} with only textual modality. The utterances in this dataset are annotated into seven classes, and they are \textit{neutral}, \textit{joyful}, \textit{peaceful}, \textit{powerful}, \textit{scared}, \textit{mad}, and \textit{sad}. 

\textbf{DailyDialog} contains 13,118 conversations with a total of 102,979 utterances. The dataset is a large-scale multi-turn dyadic dialogue dataset with the conversations reflecting various topics in daily life. Each utterance in the dataset is labeled with one of seven emotion categories: \textit{neutral}, \textit{happiness}, \textit{surprise}, \textit{sadness}, \textit{anger}, \textit{disgust}, and \textit{fear}. The dataset suffers from a severe class imbalance, with over 83\% of emotion labels being \textit{neutral}.

\begin{table}[htbp]
    \centering
    \renewcommand{\arraystretch}{1.0}
    \setlength{\tabcolsep}{3pt}
    \caption{The statistics for these four emotion datasets. Here, \#Con and \#Utt denote the number of conversations and utterances, respectively.}
    \begin{tabular}{c|c|cccc}
    \hline
    \multicolumn{2}{c|}{\textbf{Dataset}} &MELD &IEMOCAP &EmoryNLP &DailyDialog\\
    \hline
    \multirow{3}{*}{\#Con} &Train &1,039 &\multirow{2}{*}{120} &659  &11,118\\
     &Val &114 &  &89 &1,000\\
     &Test &280 &31 &79 &1,000\\
    \hline
    \multirow{3}{*}{\#Utt} &Train &9,989 &\multirow{2}{*}{5,810} &7,551 &87,170\\
     &Val &1,109 &  &954 &8,069\\
     &Test &2,610 &1,623 &984 &7,740\\
    \hline
    \end{tabular}
    \label{tab:statistics}
\end{table}
The statistics are reported in Table~\ref{tab:statistics}. According to COSMIC~\cite[]{ghosal2020cosmic}, we use utterance-level textual features which are fine-tuned adopting RoBERTa~\cite[]{liu2020roberta} to implement ERC task. Note that we don't adopt the features of other modalities (i.e., acoustic and visual modalities), and the split of training/validation/testing is consistent with COSMIC.

\subsection{Baselines}
To demonstrate the validity of the proposed ERNetCL, we choose some representative baselines for comparison, as follows: 

\textbf{COSMIC}~\cite[]{ghosal2020cosmic} leveraged multiple GRUs and incorporated commonsense knowledge to capture complex interactions. 

\textbf{HiTrans}~\cite[]{li2020hitrans} proposed a hierarchical framework that consists of two different Transformers to capture context- and speaker-sensitive information. 

\textbf{AGHMN}~\cite[]{jiao2020real} proposed a hierarchical memory network and attention GRU to better utilize the attention weights to improve the performance. 

\textbf{DialogueCRN}~\cite[]{hu2021dialoguecrn} was a cognitive-inspired network that utilized LSTM networks to construct multi-turn reasoning modules and capture implicit emotional clues in the conversation. 

\textbf{SKAIG-ERC}~\cite[]{li2021pastpresent} generated the knowledge representation of edges with the aid of commonsense knowledge to enhance emotional expression. 

\textbf{DialogXL}~\cite[]{shen2021dialogxl} utilized a modified XLNet~\cite[]{yang2019xlnet} to encode multi-turn dialogues that were organized in a sliding window, and its dialog-aware self-attention contained four distinct attention modules. 

\textbf{CauAIN}~\cite[]{zhao2022cauain} retrieved causal clues in commonsense knowledge to enrich the modeling of intra- and inter-speaker dependencies. 

\textbf{CoG-BART}~\cite[]{li2022contrast} proposed a novel ERC approach that adopted both contrastive learning and generative modeling to ensure the fact that different emotions were mutually exclusive.

\subsection{Training settings}
The operating system which we use is Ubuntu 20.04, and the programming language is Python 3.9.12. Our experiments are conducted on a single NVIDIA GeForce RTX 3090, the CUDA version is 11.7, and the deep learning framework is PyTorch with version 2.0.0. AdamW~\cite[]{loshchilov2018decoupled} is selected as the optimizer, and the L2 regularization factor is 3e-4. The random seed is set to 2023, and the number of heads in each MHA network is set to 4. Table~\ref{tab:hyperparameter} describes the partial hyperparameter settings for different datasets.
\begin{table}[htbp]
    \centering
    \renewcommand{\arraystretch}{1.0}
    \setlength{\tabcolsep}{4pt}
    \caption{Partial hyperparameter settings. Here, LR, BS, and DR denote the learning rate, batch size, and dropout rate, respectively; DepthT and DepthS indicate the depth of temporal encoder and spatial encoder, respectively.}
    \begin{tabular}{c|ccccccc}
    \hline
    \textbf{Dataset} &LR &$\delta$ &$\sigma$ &BS &DepthT &DepthS &DR \\
    \hline
    MELD &1e-5 &10 &0.4 &128 &4 &4 &0.2 \\
    IEMOCAP &1e-4 &9 &0.7 &64 &2 &6 &0.1 \\
    EmoryNLP &1e-5 &7 &0.6 &128 &4 &3 &0.2 \\
    DailyDialog &1e-5 &9 &0.6 &128 &3 &6 &0.2 \\
    \hline
    \end{tabular}
    \label{tab:hyperparameter}
\end{table}

\section{Results and analysis}\label{sec:result}
In this section, we report the experimental results of the proposed ERNetCL on these four datasets. We compare these results with those of the baselines. For the MELD, IEMOCAP, and EmoryNLP datasets, we utilize weighted F1 and micro F1 scores as evaluation metrics. For the DailyDialog dataset, we adopt macro F1 score and micro F1 score without \textit{neutral} to evaluate our model since \textit{neutral} accounts for about 83\% in the dataset. In addition, we investigate the impact of different modules or settings on the performance of ERNetCL.

\subsection{Overall comparison}
\begin{table*}[htbp]
    \centering
    \renewcommand{\arraystretch}{1.0}
    \setlength{\tabcolsep}{6.0pt}
    \caption{Performance comparison of ERNetCL with all baselines. Results for all baselines are obtained from the original paper. Best results are highlighted in bold.}
    \begin{tabular}{c|cc|cc|cc|cc}
    \hline
    \multirow{2}{*}{\textbf{Method}} &\multicolumn{2}{c|}{MELD} &\multicolumn{2}{c|}{IEMOCAP} &\multicolumn{2}{c|}{EmoryNLP} &\multicolumn{2}{c}{DailyDialog}\\
    \cline{2-9}
           &weighted-F1 &micro-F1 &weighted-F1 &micro-F1 &weighted-F1 &micro-F1 &macro-F1 &micro-F1\\ 
    \hline
    COSMIC~\cite[]{ghosal2020cosmic} &65.21 &- &65.28 &- &38.11 &-  &51.05 &58.48\\
    HiTrans~\cite[]{li2020hitrans} &61.94 &- &64.50 &- &36.75 &- &- &- \\
    AGHMN~\cite[]{jiao2020real} &58.10 &- &63.50 &- &- &- &- &- \\
    DialogueCRN~\cite[]{hu2021dialoguecrn} &58.39 &- &66.20 &- &- &- &- &- \\
    SKAIG-ERC~\cite[]{li2021pastpresent} &65.18 &- &66.96 &- &38.88 &- &51.95 &59.75 \\
    DialogXL~\cite[]{shen2021dialogxl} &62.41 &- &65.94 &- &34.73 &-  &- &54.93\\
    CauAIN~\cite[]{zhao2022cauain} &65.46 &- &67.61 &- &- &-  &\textbf{53.85} &58.21 \\
    CoG-BART~\cite[]{li2022contrast} &64.81 &65.95 &66.18 &66.71 &39.04 &42.58 &- &56.29 \\
    \hline
    ERNetCL &\textbf{66.31} &\textbf{67.43} &\textbf{69.73} &\textbf{69.75} &\textbf{39.71} &\textbf{44.21} &53.09 &\textbf{60.17}\\
    \hline
    \end{tabular}
    \label{tab:overallcomparison}
\end{table*}
The experimental results of ERNetCL and comparative models are reported in Table~\ref{tab:overallcomparison}. From these results, we can draw the following conclusions:
\begin{enumerate}[(1)]
    \item On the MELD dataset, ERNetCL attains a weighted F1 score of 66.31\%, which is 0.85\% higher than that of 65.46\% obtained by CauAIN. This suggests that even though CauAIN introduces causal cues, it still does not thoroughly model context. The proposed method achieves a micro F1 score of 67.43\%, which is 1.48\% higher relative to that of CoG-BART. A probable explanation for the discrepancy is the additional noise introduced by the response generation module in CoG-BART.
    \item On the IEMOCAP dataset, the weighted F1 score and micro F1 score of our ERNetCL are 69.73\% and 69.75\%, respectively, which are much higher than those of other models. For example, ERNetCL's weighted F1 score is 3.79\% higher than DialogXL's, while its micro F1 score improves by 3.04\% with respect to CoG-BART's. These superior results are facilitated by the fact that ERNetCL can incorporate the idea of curriculum learning and extract contextual information from both temporal and spatial perspectives.
    \item On the EmoryNLP dataset, the weighted F1 score of HiTrans is 36.75\%, which is 2.96\% lower than that of ERNetCL's 39.71\%; and the micro F1 score of CoG-BART is 42.58\%, while the score of our method is 44.21\%, which is an improvement of 1.63\%. One possible reason is that HiTrans and CoG-BART have difficulty extracting chronological information from the conversation.
    \item On the DailyDialog dataset, the macro F1 score of our ERNetCL is 53.09\%, which is 0.76\% lower than that of CauAIN; but the micro F1 score of CauAIN is 58.21\%, which is 1.96\% lower than that of our method, which is 60.05\%. The main reason for the limited performance of ERNetCL is due to the class imbalance problem in the DailyDialog dataset.
\end{enumerate}

In most cases, our model achieves optimal performance on these four datasets compared to all baselines. This indicates that our proposed ERNetCL can utilize simple encoding structures to effectively capture temporal and spatial contextual information in the conversation.

\subsection{Results for each emotion}
\begin{figure*}[htbp]
    \centering
    \subfloat[MELD dataset]{\includegraphics[height=2.4in]{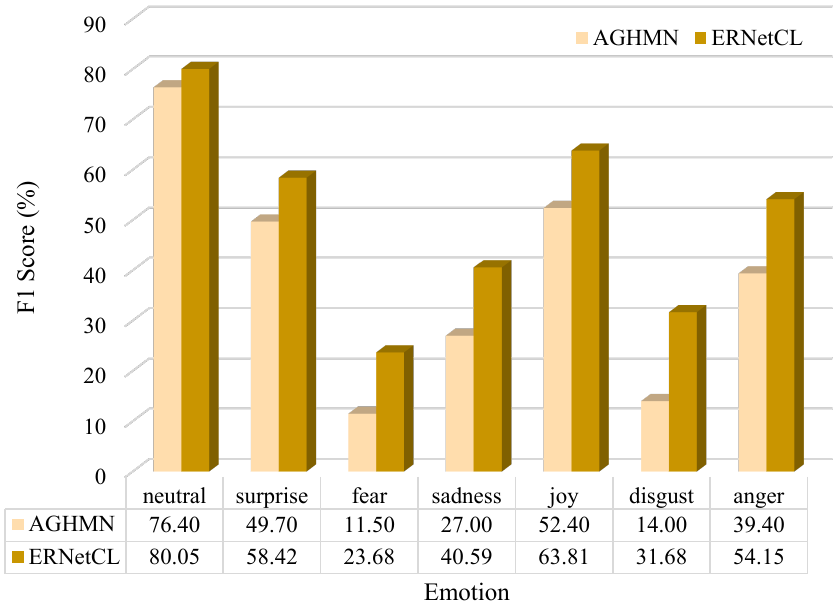}%
    \label{fig:meld_each_emotion}}
    \hfil
    \subfloat[IEMOCAP dataset]{\includegraphics[height=2.4in]{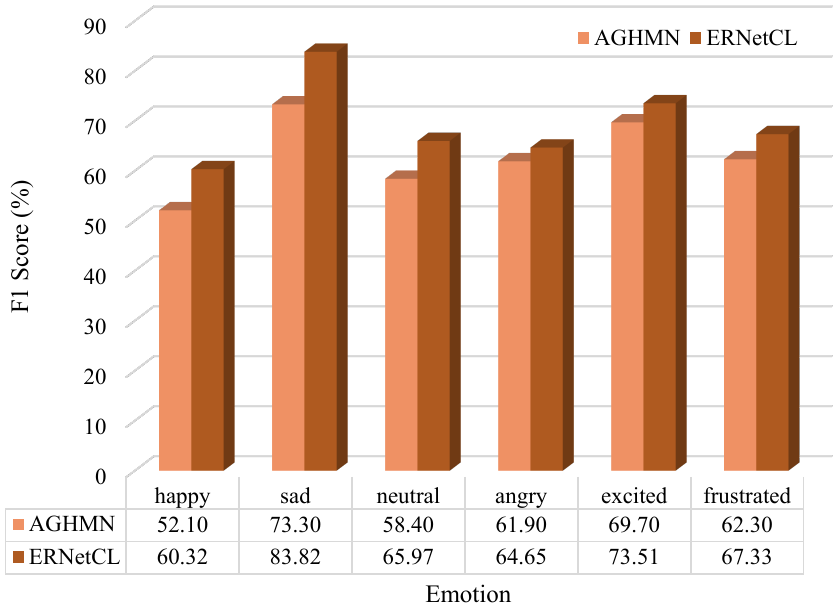}%
    \label{fig:iemocap_each_emotion}}
    \caption{F1 score for each emotion on the MELD and IEMOCAP datasets. Our model's F1 scores for all emotions are higher than AGHMN's results.}
    \label{fig:each_emotion}
\end{figure*}
As shown in Figure~\ref{fig:each_emotion}, we show the F1 scores for each emotion on the MELD and IEMOCAP datasets. On the MELD dataset, ERNetCL's F1 scores for all emotions are higher than AGHMN's results. For instance, the result of ERNetCL for \textit{surprise} is 58.42\%, which is 8.72\% higher than that of AGHMN. It is worth noting that ERNetCL's F1 scores for the minority class \textit{fear} and \textit{disgust} are significantly higher than AGHMN's results. A possible reason is that the curriculum learning loss of ERNetCL mitigates the class-imbalanced problem of the MELD dataset to some extent. On the IEMOCAP dataset, the scores of our ERNetCL for all emotions are higher than those of AGHMN. For emotion \textit{sad}, the F1 score of AGHMN is 73.30\%, while that of ERNetCL is 83.82\%, which is a significant improvement of 10.52\%.

In addition, on the MELD dataset, ERNetCL obtains the highest F1 score for \textit{neutral} in comparison to those for other emotions. By examining the class distribution of MELD, we find that \textit{neutral} belongs to the majority class by an absolute proportional advantage with a share of about 46.95\%. The emotion that achieves the highest F1 score on the IEMOCAP dataset is \textit{sad}. However, \textit{sad} is not the category with the highest percentage in the IEMOCAP dataset. The main reason for this phenomenon is that the class distribution of IEMOCAP is relatively balanced, and the slight class-imbalanced problem barely affects the performance of the model.

\begin{figure}[htbp]
    \centering
    \subfloat[Initial Features]{\includegraphics[height=1.6in]{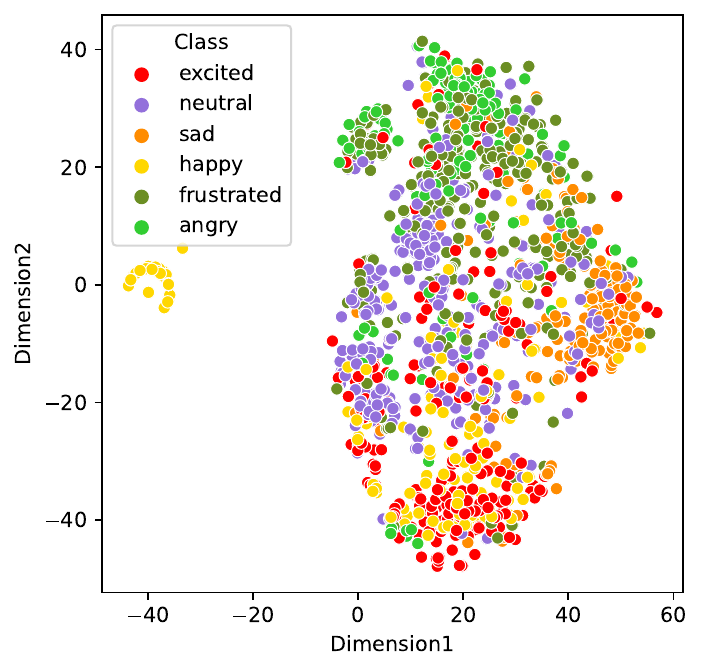}%
    \label{fig:iemocap_initial_emo}}
    \hfil
    \subfloat[Extracted Features]{\includegraphics[height=1.6in]{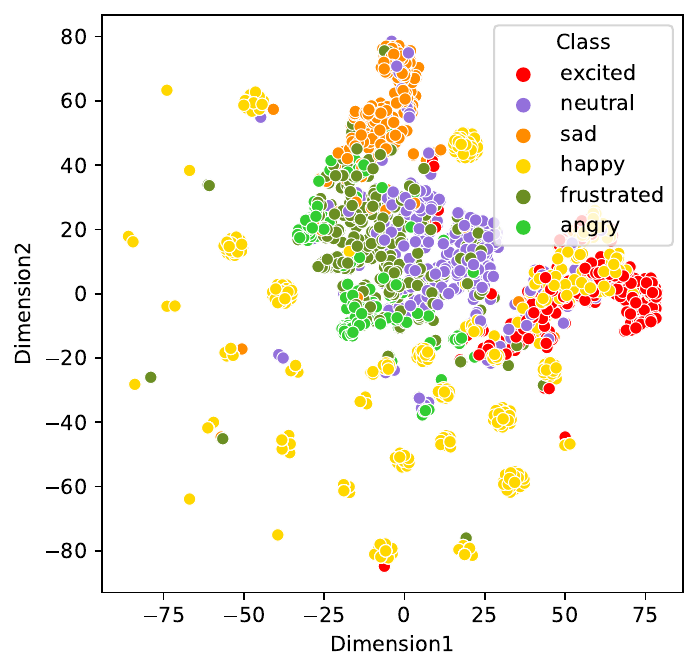}%
    \label{fig:iemocap_cfnesa_emo}}
    \caption{T-SNE visualization of IEMOCAP before and after feature extraction.}
    \label{fig:iemocap_tsne_emo}
\end{figure}

Figure~\ref{fig:iemocap_tsne_emo} shows the T-SNE visualization of the IEMOCAP dataset. It can be seen that each emotion can be distinguished more easily after feature extraction with the proposed ERNetCL. This situation shows that ERNetCL is powerful in feature extraction.

\subsection{Ablation studies}
\begin{table}[htbp]
    \centering
    \renewcommand{\arraystretch}{1.0}
    \setlength{\tabcolsep}{3pt}
    \caption{Results after removing each component on the MELD and IEMOCAP datasets. Here, -w/o TE, -w/o SE, and -w/o CL denote the removal of the temporal encoder, spatial encoder, and curriculum learning loss (replacing with the standard cross-entropy loss), respectively.}
    \begin{tabular}{c|cc|cc}
    \hline
    \multirow{2}{*}{\textbf{Method}} &\multicolumn{2}{c|}{MELD} &\multicolumn{2}{c}{IEMOCAP}\\
           &weighted-F1 &micro-F1 &weighted-F1 &micro-F1\\ 
    \hline 
	ERNetCL &\textbf{66.31} &\textbf{67.43} &\textbf{69.73} &\textbf{69.75} \\
	\hline
	-w/o TE &64.47 &65.86 &65.89 &65.99 \\
    -w/o SE &65.98 &66.97 &65.58 &65.87 \\
	-w/o CL &65.98 &67.01 &67.84 &67.96 \\
	\hline
    \end{tabular}
    \label{tab:ablation}
\end{table}
\begin{figure}[htbp]
    \centering
    \includegraphics[width=3.3in]{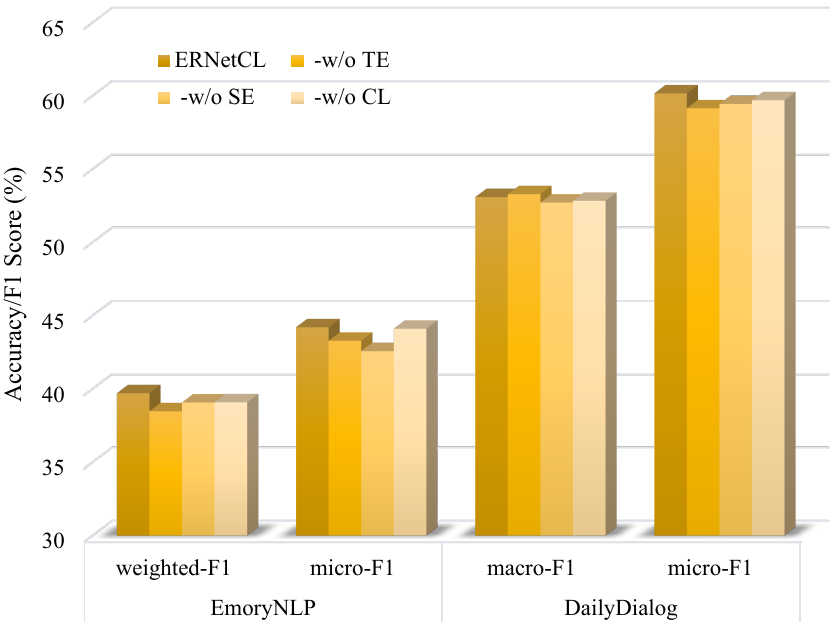}
    \caption{Results after removing each component on the EmoryNLP and DailyDialog datasets.}
    \label{fig:ablation}
\end{figure}
In order to explore the contribution of each component (i.e., SE, TE, and CL) separately, we report the experimental results regarding the removal of different components in Table~\ref{tab:ablation} and Figure~\ref{fig:ablation}. Overall, regardless of which component is removed, the performance of ERNetCL suffers a degradation.

Effectiveness of TE: Table~\ref{tab:ablation} shows the results on the MELD and IEMOCAP datasets. When TE is removed, the weighted F1 scores of our model on these two datasets decrease by 1.84\% and 3.84\%, respectively. As shown in Figure~\ref{fig:ablation}, on the EmoryNLP and DailyDialog datasets, removing TE also leads to the performance degradation of the model. These phenomenons suggest that TE can effectively capture temporal contextual information.

Effectiveness of SE: As can be seen in Table~\ref{tab:ablation}, when we remove SE, there are decreases of 0.46\% and 3.88\% in the micro F1 scores on the MELD and IEMOCAP datasets, respectively. It can also be observed from Figure~\ref{fig:ablation} that the similar declines occur on the EmoryNLP and DailyDialog datasets when SE is removed. The above results indicate that SE can effectively extract spatial context information.

Effectiveness of CL: We remove the CL loss and replace it with the standard cross-entropy loss (i.e., Equation~\ref{eq:loss_}). As shown in Table~\ref{tab:ablation}, non-use of the CL loss brings about 0.42\% and 1.79\% drops in micro F1 scores on the MELD and IEMOCAP datasets, respectively. A similar situation appears on the other two datasets, as shown in Figure~\ref{fig:ablation}. The above phenomenon reveals that CL can reduce the learning difficulty of the network and effectively promote the performance of the model. Thus, it can be concluded that CL loss prevails over the standard loss and somewhat eases the detrimental impacts associated with emotion shift.

\subsection{Impact of network depths}
\begin{figure}[htbp]
    \centering
    \includegraphics[width=3.3in]{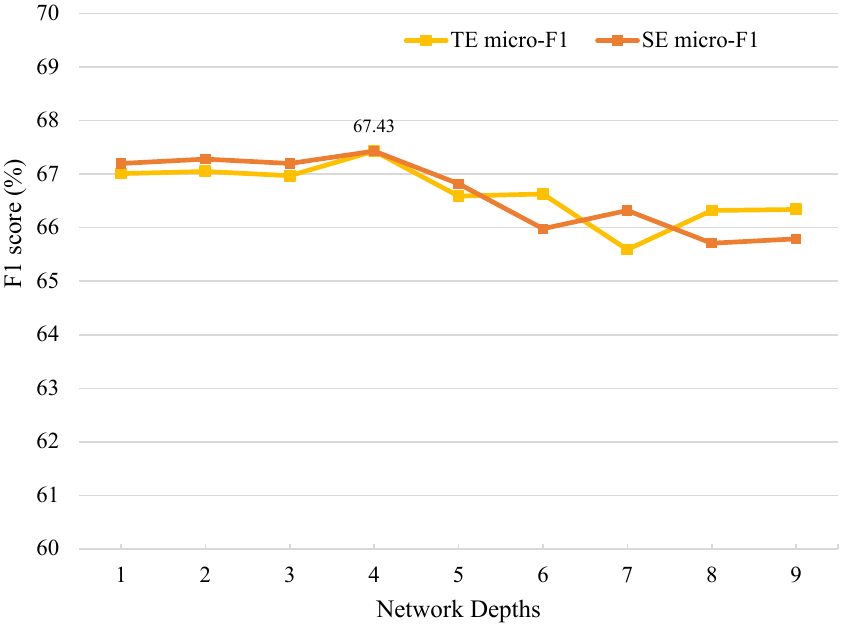}
    \caption{The impact of the network depths on the performance. Experimental results on the MELD dataset. The yellow and orange lines indicate the effect of the temporal and spatial encoders on the performance, respectively.}
    \label{fig:networkdepths}
\end{figure}
To examine the impact of TE with different network depths (number of layers) on the performance, we first fix the network depth of SE, then adjust that of TE and report the micro F1 scores. As shown in Figure~\ref{fig:networkdepths}, the yellow line is the experimental result for testing the network depths of TE. It can be viewed that as the network depth increases, the micro F1 score shows a tendency to increase and then decrease. Similarly, by fixing the network depth of TE, we can explore the effect of SE with different network depths on the performance. The experimental results are shown with the orange line in Figure~\ref{fig:networkdepths}, where the performance increases and then decreases as the number of network layers grows.

\subsection{Results of sentiment classification}
\begin{table}[htbp]
    \centering
    \renewcommand{\arraystretch}{1.0}
    \setlength{\tabcolsep}{2pt}
    \caption{Comparison for emotion and sentiment classification. Results for COSMIC are obtained from the original paper. Here, -E and -S denote the emotion and sentiment classification, respectively.}
    \begin{tabular}{c|cc|cc}
    \hline
    \multirow{2}{*}{\textbf{Method}} &\multicolumn{2}{c|}{MELD} &\multicolumn{2}{c}{EmoryNLP}\\
    \cline{2-5}
        &weighted-F1 &micro-F1 &weighted-F1 &micro-F1\\ 
    \hline
    COSMIC-E &65.21 &- &38.11 &-  \\
    COSMIC-S &73.20 &- &56.51 &-  \\
    \hline
    ERNetCL-E &{66.31} &{67.43} &{39.71} &{44.21}  \\
    ERNetCL-S &{73.27} &{73.26} &{57.06} &{57.11}  \\
    \hline
    \end{tabular}
    \label{tab:sentiment}
\end{table}
\begin{figure}[htbp]
    \centering
    \subfloat[Emotion Classification]{\includegraphics[height=1.6in]{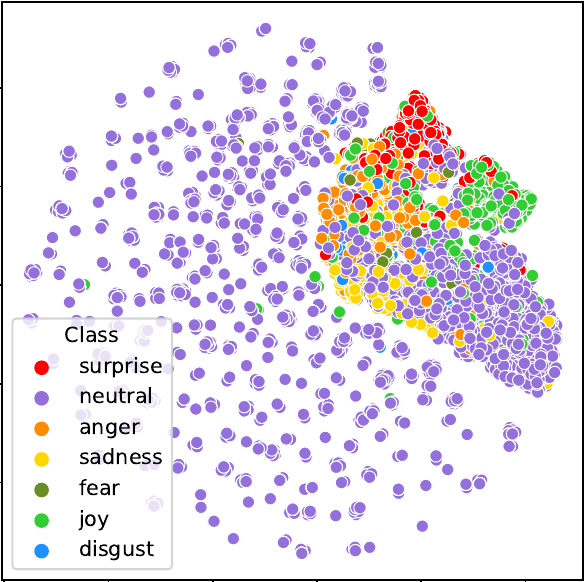}%
    \label{fig:meld_ernetcl_emo}}
    \hfil
    \subfloat[Sentiment Classification]{\includegraphics[height=1.6in]{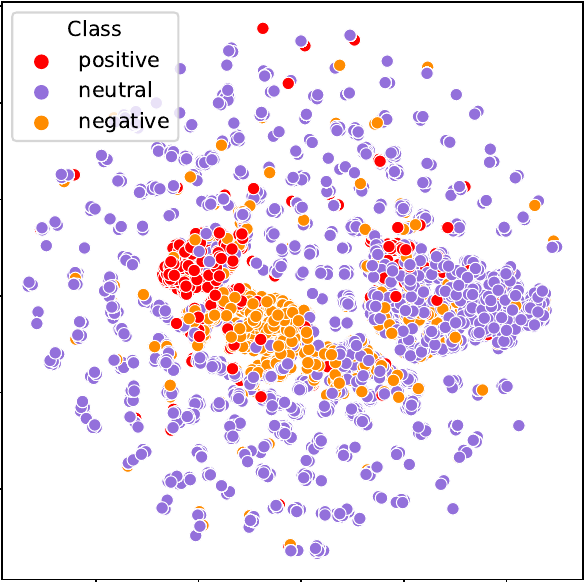}%
    \label{fig:meld_ernetcl_sent}}
    \caption{T-SNE visualization of emotion and sentiment classification on the MELD dataset.}
    \label{fig:meld_tsne}
\end{figure}
Instead of using emotion labels, we use sentiment labels for the three-classification task. As can be seen in Table~\ref{tab:sentiment}, all the experimental results show significant improvements relative to those for emotion classification. For example, the weighted F1 scores on the MELD and EmoryNLP datasets are improved by 6.96\% and 17.35\%, respectively. This is due to the fact that the classification difficulty is reduced after the emotion labels are coarsened to sentiments. This phenomenon can be illustrated with the T-SNE visualization in Figure~\ref{fig:meld_tsne}. The sentiment classification is easier to distinguish each category than the emotion classification. Compared to the weighted F1 scores of COSMIC under sentiment classification, our scores are only marginally improved. A possible explanation is that the lower the difficulty of the dataset, the less the feature extraction capability is required of the model. Thus, ERNetCL-S and COSMIC-S have similar results.

\subsection{Limitations}
\begin{figure*}[htbp]
    \centering
    \subfloat[MELD Dataset]{\includegraphics[height=2.5in]{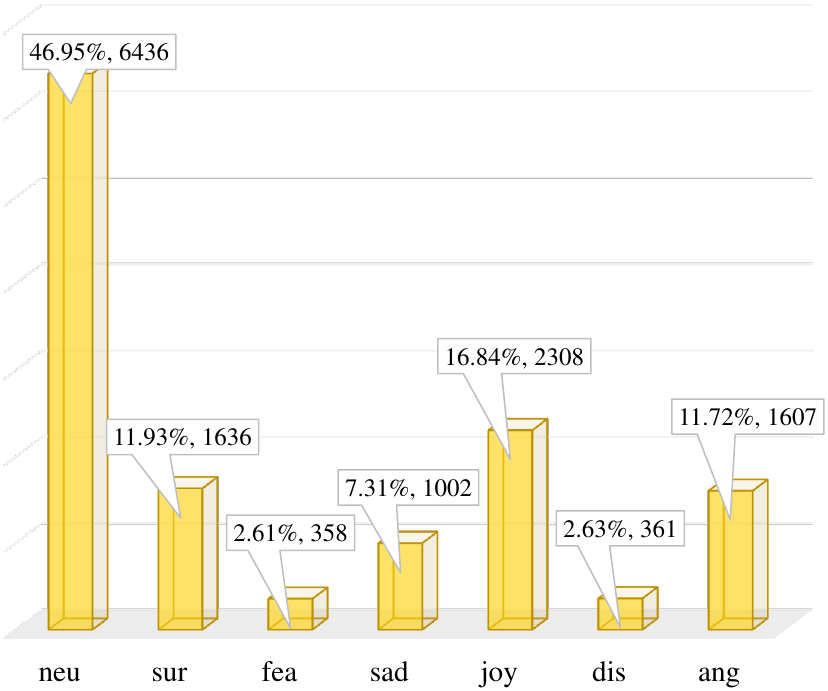}%
    \label{fig:meld_utter}}
    \hfil
    \subfloat[DailyDialog Dataset]{\includegraphics[height=2.5in]{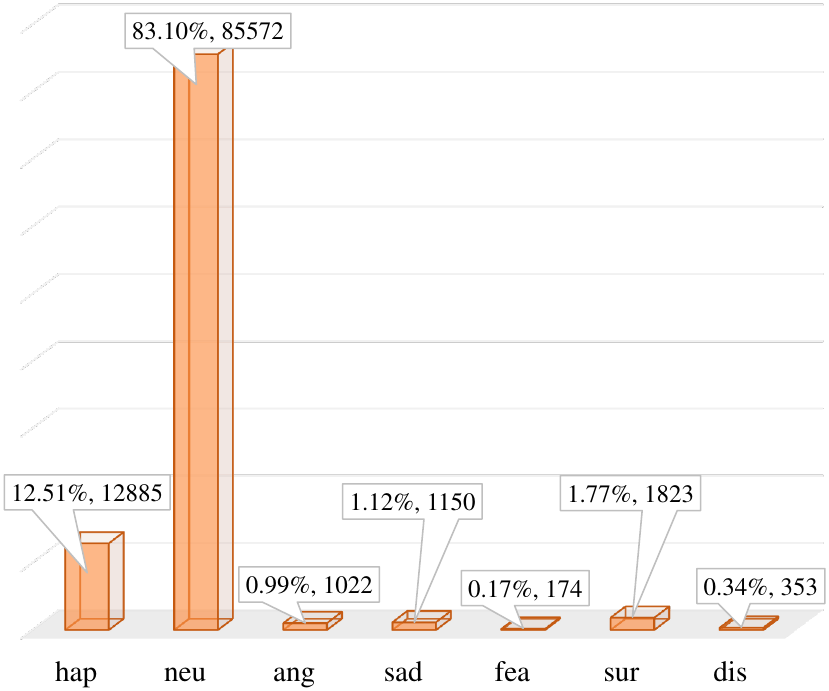}%
    \label{fig:dailydialog_utter}}
    \caption{The number of utterances for each emotion in the MELD and DailyDialog datasets. The data annotation includes proportion and quantity. \textit{neu} represents for the first three-letter abbreviation for \textit{neutral}, and so on for other emotions.}
    \label{fig:utter}
\end{figure*}
\begin{figure*}[htbp]
    \centering
    \subfloat[MELD Dataset]{\includegraphics[height=2.5in]{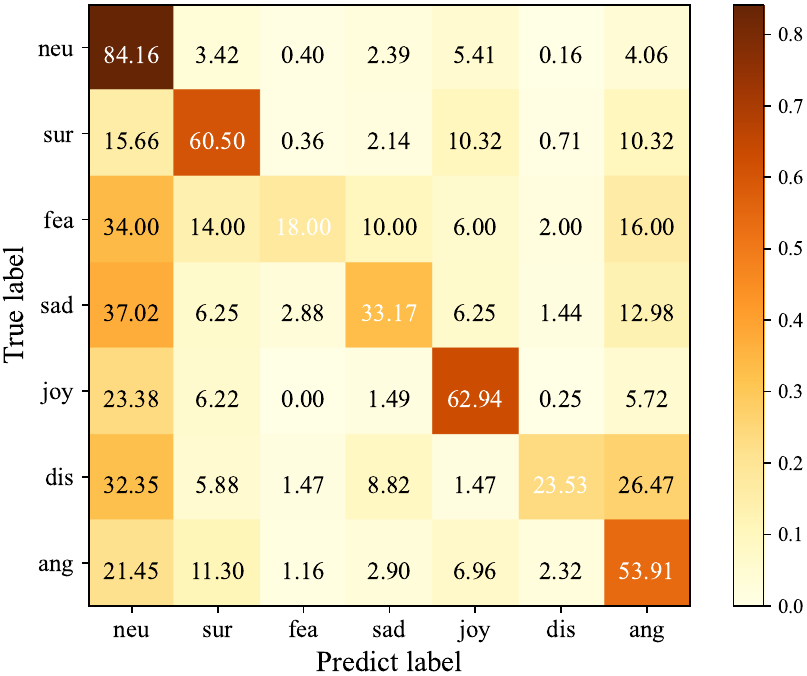}%
    \label{fig:meld_cm}}
    \hfil
    \subfloat[DailyDialog Dataset]{\includegraphics[height=2.5in]{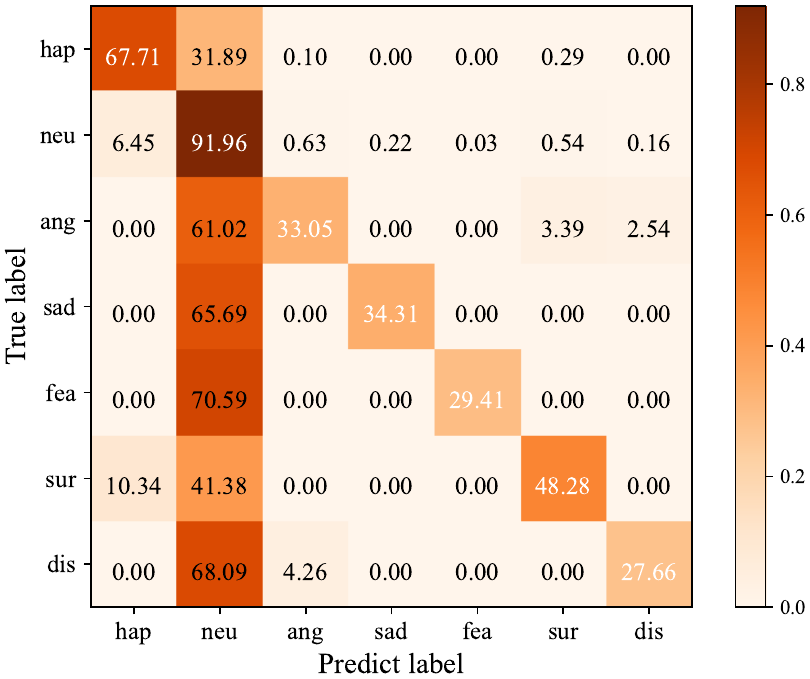}%
    \label{fig:dailydialog_cm}}
    \caption{The confusion matrices on the MELD and DailyDialog datasets. The vertical and horizontal axes denote the true and predicted labels, respectively.}
    \label{fig:cm}
\end{figure*}
The categories in most emotion datasets are unbalanced with each other, i.e., there is a class-imbalanced problem. Figure~\ref{fig:utter} illustrates the number of utterances for each emotion class in the MELD and DailyDialog datasets. It can be seen that in the MELD dataset, \textit{neutral} is the majority class with about 47\%; in the DailyDialog dataset, \textit{neutral} is the majority class with approximately 83\%. Like previous ERC models, the performance of ERNetCL is limited by the class-imbalanced problem. Figure~\ref{fig:cm} shows confusion matrices on the MELD and DailyDialog datasets. In the MELD dataset, three minority classes, i.e., \textit{fear}, \textit{sad}, and \textit{disgust}, are easily recognized as the majority class \textit{neutral}. Since DailyDialog is an extremely class-imbalanced dataset, most emotions tend to be detected as the majority class \textit{neutral}. Additionally, another limitation is the similar-emotional problem, that is, an emotion class is easily categorized as the similar emotion in some cases. As shown in Figure~\ref{fig:meld_cm}, the true emotion \textit{disgust} is easily recognized as \textit{anger}.

\section{Conclusion}\label{sec:conclusion}
Previous ERC approaches fail to model the context from both temporal and spatial perspectives, and the network designs of these models are overly complex, e.g., requiring commonsense knowledge, employing encoder-decoder architecture, and containing too many components. The above issues may cause these models to generate redundant information, leading to weak performance enhancements. In this work, we propose a novel emotion recognition method, ERNetCL, to fully extract contextual emotional cues from the conversation. ERNetCL combines RNN- and MHA-based approaches straightforwardly, employing temporal and spatial encoders to capture the contextual information of the utterance from the temporal and spatial perspectives, respectively. Furthermore, to relieve the adverse affects caused by emotion shift and further enhance the performance of the model, we employ the CL strategy to train the proposed model in a fashion that simulates humans learning curriculum. We define the difficulty score of CL using the frequency of emotion shifts in the conversation, assigning different learning weights to the samples in the training set. We conduct extensive experiments on four widely used datasets to evaluate the proposed method. Empirical results attest that ERNetCL can effectively model context in a simple manner and significantly outperforms other comparative models in most cases. 

In future work, we will investigate the ERC tasks based on multimodal fusion and explore the further application of curriculum learning in them. Moreover, reinforcement learning and contrast learning have recently received a great deal of attention. The former can utilize the reward mechanism to return a higher score for the minority class and a lower score for the majority class. Contrastive learning can pull the same classes closer together and push different classes farther apart. Therefore, we intend to integrate these techniques into the ERC model to overcome the class-imbalanced and similar-emotional problems.

\section*{CRediT authorship contribution statement}
\textbf{Jiang Li:} Conceptualization, Methodology, Data curation, Software, Validation, Formal analysis, Investigation, Visualization, Writing - original draft, Writing - review \& editing, Project administration. \textbf{Xiaoping Wang:} Supervision, Writing - review \& editing, Funding acquisition. \textbf{Yingjian Liu:} Conceptualization, Writing - review \& editing. \textbf{Zhigang Zeng:} Supervision, Funding acquisition.

\section*{Declaration of competing interest}
The authors declare that they have no known competing financial interests or personal relationships that could have appeared to influence the work reported in this paper.

\section*{Acknowledgments}
This work was supported in part by the National Natural Science Foundation of China under Grant 62236005, 61936004, and U1913602. We thank the reviewers and editorial board for helpful comments that greatly improved the paper.

\bibliographystyle{elsarticle-num}
\bibliography{ernetcl.bib}
\balance

\end{document}